\documentclass[letterpaper, 10 pt, journal, twoside]{IEEEtran}  % Comment this line out if you need a4paper

\newcommand{\subparagraph}{}

\usepackage{subcaption}
\usepackage{graphicx}
\usepackage{algpseudocode}
\usepackage{algorithm}
\usepackage{amsmath}
\usepackage{verbatim}
\usepackage{amsfonts}
\usepackage{amssymb}
\usepackage{placeins}
\usepackage[normalem]{ulem}
\usepackage{enumerate}
\usepackage{color, soul}
\usepackage{mathtools}
\usepackage{array,arydshln}
\usepackage{url}
\usepackage{titlesec}

\algnewcommand\algorithmicforeach{\textbf{for each}}
\algdef{S}[FOR]{ForEach}[1]{\algorithmicforeach\ #1\ \algorithmicdo}
\DeclareMathOperator*{\argmax}{argmax}
\DeclareMathOperator*{\argmin}{argmin}

\newtheorem{remark}{Remark}
\newtheorem{theorem}{Theorem}

 \author{Dimitrios I. Koutras$^{1}$, Athanasios Ch. Kapoutsis$^{2}$ and Elias B. Kosmatopoulos$^{1}$
 \thanks{Manuscript received: February, 24, 2020; Revised May, 20, 2020; Accepted June, 8, 2020.}
 \thanks{This paper was recommended for publication by Editor Jonathan Roberts upon evaluation of the Associate Editor and Reviewers' comments. This project has received funding from the European Commission under the European Union's Horizon 2020 research and innovation programme under grant agreement no 833464 (CREST). (\textit{Corresponding author: Athanasios Ch. Kapoutsis})}
	\thanks{$^{1}$ Dimitrios I. Koutras and Elias B. Kosmatopoulos are with Department of Electrical and Computer Engineering,    Democritus University of Thrace, Xanthi, Greece and Information Technologies Institute, The Centre for Research \& Technology, Hellas, Thessaloniki, Greece (dkoutras@iti.gr, kosmatop@iti.gr)}%
	\thanks{$^{2}$Athanasios Ch. Kapoutsis is with Information Technologies Institute, The Centre for Research \& Technology, Hellas, Thessaloniki, Greece (athakapo@iti.gr) } %{\tt\small athakapo@iti.gr}
 \thanks{Digital Object Identifier (DOI): 10.1109/LRA.2020.3004780}
 } %Use only for final RAL version.

\title{Autonomous and cooperative design of the monitor positions for a team of UAVs to maximize the quantity and quality of detected objects}
\begin{document}

	\maketitle
	%\thispagestyle{empty}
	%\pagestyle{empty}

	%%%%%%%%%%%%%%%%%%%%%%%%%%%%%%%%%%%%%%%%%%%%%%%%%%%%%%%%%%%%%%%%%%%%%%%%%%%%%%%%
	\begin{abstract}
		This paper tackles the problem of positioning a swarm of UAVs inside a completely unknown terrain, having as objective to maximize the overall situational awareness. The situational awareness is expressed by the number and quality of unique objects of interest, inside the UAVs' fields of view. YOLOv3 and a system to identify duplicate objects of interest were employed to assign a single score to each UAVs' configuration. Then, a novel navigation algorithm, capable of optimizing the previously defined score, without taking into consideration the dynamics of either UAVs or environment, is proposed. A cornerstone of the proposed approach is that it shares the same convergence characteristics as the block coordinate descent (BCD) family of approaches. The effectiveness and performance of the proposed navigation scheme were evaluated utilizing a series of experiments inside the AirSim simulator. The experimental evaluation indicates that the proposed navigation algorithm was able to consistently navigate the swarm of UAVs to ``strategic'' monitoring positions and also adapt to the different number of swarm sizes. Source code is available at \url{https://github.com/dimikout3/ConvCAOAirSim}.
	\end{abstract}
	%%%%%%%%%%%%%%%%%%%%%%%%%%%%%%%%%%%%%%%%%%%%%%%%%%%%%%%%%%%%%%%%%%%%%%%%%%%%%%%%
	
	\begin{IEEEkeywords}
		Perception and Autonomy, Motion and Path Planning
	\end{IEEEkeywords}

	%\markboth{IEEE Robotics and Automation Letters. Preprint Version. Accepted June, 2020}
	
	\section{INTRODUCTION}
	\label{sec:intro}
	
	\begin{figure*}
		\centering
		\begin{subfigure}[b]{0.465\textwidth}
			\begin{subfigure}[b]{1\textwidth}
				\includegraphics[width=\textwidth]{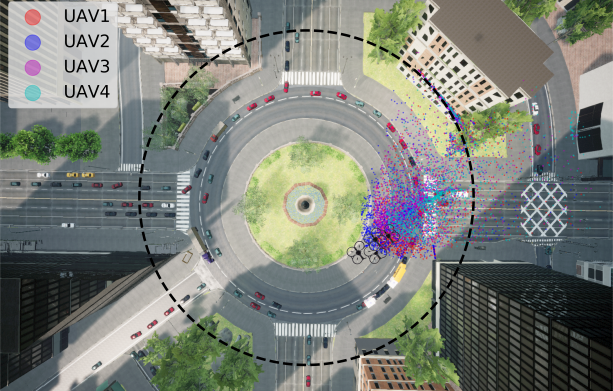}
				%\caption*{Bird's eye view}
				\label{fig:globalViewInitial}
			\end{subfigure}
			\begin{subfigure}[b]{1\textwidth}
				\begin{subfigure}[b]{0.475\textwidth}
					\includegraphics[width=1\textwidth]{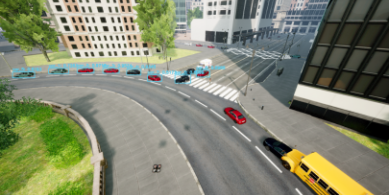}
					\caption*{UAV1: 5 \textit{unique} out of 13 total detected objects}
					\label{fig:UAV1Initial}
				\end{subfigure}
				\begin{subfigure}[b]{0.475\textwidth}
					\includegraphics[width=1\textwidth]{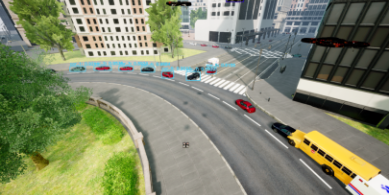}
					\caption*{UAV2: 2 \textit{unique} out of 5 total detected objects}
					\label{fig:UAV2Initial}
				\end{subfigure}
				\begin{subfigure}[b]{0.475\textwidth}
					\includegraphics[width=1\textwidth]{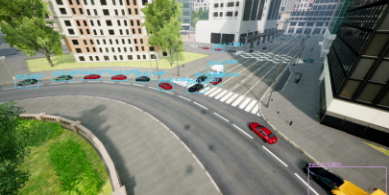}
					\caption*{UAV3: 2 \textit{unique} out of 7 total detected objects}
					\label{fig:UAV3Initial}
				\end{subfigure}
				\;\;
				\begin{subfigure}[b]{0.475\textwidth}
					\includegraphics[width=1\textwidth]{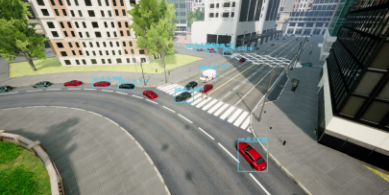}
					\caption*{UAV4: 4 \textit{unique} out of 9 total detected objects}
					\label{fig:UAV4Initial}
				\end{subfigure}
			\end{subfigure}
			\caption{Initial monitor positions}
			\label{fig:introInitial}
		\end{subfigure}
		%\quad \qquad
		%\rulesep
		%\hfill\vline\hfill
		%\hfill\vline\hfill
		~ %add desired spacing between images, e. g. ~, \quad, \qquad, \hfill etc. 
		~ %add desired spacing between images, e. g. ~, \quad, \qquad, \hfill etc. 
		\begin{subfigure}[b]{0.465\textwidth}
			\begin{subfigure}[b]{1\textwidth}
				\includegraphics[width=1\textwidth]{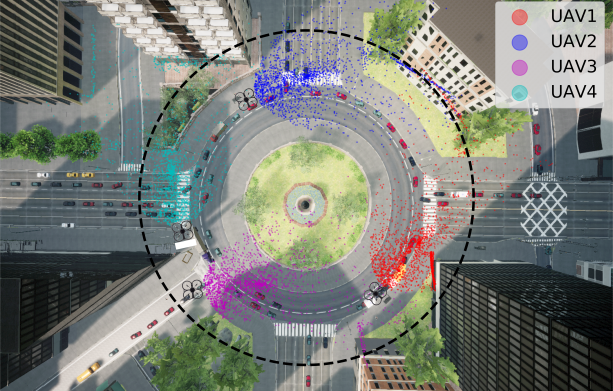}
				%\caption*{Bird's eye view}
				\label{fig:finalGlobalView}
			\end{subfigure}
			\begin{subfigure}[b]{1\textwidth}
				\begin{subfigure}[b]{0.475\textwidth}
					\includegraphics[width=1\textwidth]{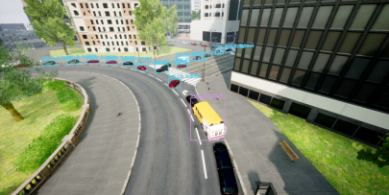}
					\caption*{UAV1: 12 \textit{unique} out of 14 total detected objects}
					\label{fig:UAV1Final}
				\end{subfigure}
				\begin{subfigure}[b]{0.475\textwidth}
					\includegraphics[width=1\textwidth]{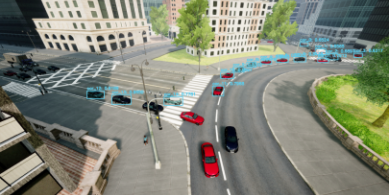}
					\caption*{UAV2: 14 \textit{unique} out of 17 total detected objects}
					\label{fig:UAV2Final}
				\end{subfigure}
				\begin{subfigure}[b]{0.475\textwidth}
					\includegraphics[width=1\textwidth]{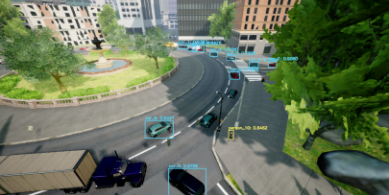}
					\caption*{UAV3: 7 \textit{unique} out of 9 total detected objects}
					\label{fig:UAV3Final}
				\end{subfigure} 
				\;\;
				\begin{subfigure}[b]{0.475\textwidth}
					\includegraphics[width=1\textwidth]{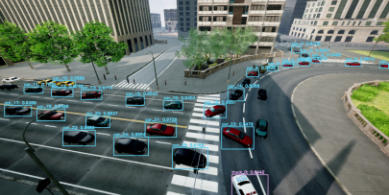}
					\caption*{UAV4: 14 \textit{unique} out of 23 total detected objects}
					\label{fig:UAV4Final}
				\end{subfigure}
			\end{subfigure}
			\caption{Monitor positions as calculated by the proposed algorithm.}
			\label{fig:introFinal}
		\end{subfigure}
		\caption{Illustrative example: A swarm of 4 UAVs is deployed above the depicted roundabout, having as objective to find the monitoring poses (position \& orientation) inside the operational area (black, dashed circle), that maximize the overall situational awareness. In both instances [sub-figures (a) \& (b)], the fields of view (point-clouds) of all the UAVs (top bird's-eye view) and also separate views - one for each UAV - with the detected objects (cyan bounding boxes) are depicted.}
		\label{fig:introFig}
	\end{figure*}

	%\cite{deng2013recent}
	
	\IEEEPARstart{T}he recent technological advancements in terms of hardware \cite{sarbazi2016advances} have enabled the development of deep learning architectures \cite{lecun2015deep} capable of dealing with extremely difficult \cite{mnih2013playing} and abstract \cite{Silver1140} problems. One of the most well-studied problems is the task of training a deep neural network to detect objects of interest \cite{han2018advanced}. The maturity of studies and available tools has reached an unprecedented level, having approaches focusing on accuracy \cite{ren2015faster}, speed \cite{yolov3}, exploiting problem-specific characteristics to boost the performance \cite{lin2017feature}, etc. The concurrent developments in the field of UAVs have allowed the introduction of these methodologies to aerial images \cite{xia2018dota} and also the development of special-purpose algorithms to detect and/or track objects of interest with one \cite{bazi2018convolutional} or more UAVs \cite{gu2018multiple}.
	
	In this paper, we tackle the same problem but from a different perspective. Instead of trying to design another object detection algorithm, we attempt to optimize the monitor positions of the swarm of the UAVs that, for a given object detection mechanism, will result in the best possible outcome. In a nutshell, we seek to answer the question:  How can we ``steer''/guide a team of UAVs to increase the number and accuracy of detected objects that lie on their combined field of view? 
	
	This task can be of paramount importance in scenarios where is needed: i) automatic surveillance of an operational area \cite{doitsidis20113d}, ii) increased situational awareness with limited resources \cite{hussein2012aegis}, iii) automatic readjustment in response to changes on the environment \cite{kapoutsis2019distributed}, etc. The aforementioned scenarios can be directly utilized in robotic applications such as: traffic monitoring \cite{kamijo2000traffic}, border surveillance \cite{girard2004border}, supervision on large crowds \cite{motlagh2017uav}, search and rescue \cite{scherer2015autonomous} etc.
	
	The vast majority of the available methodologies deals with the problem of detecting objects from given aerial captures \cite{schumann2017deep, xia2018dota}. Although many methodologies are capable of detecting the objects of interest in real-time, most of them rely on either predefined or human-controlled paths for the UAVs \cite{tang2017fast, bazi2018convolutional}. Without active feedback, regarding the progress of the detection task, the coordination among different UAVs is usually achieved by assigning spatially exclusive areas to each member of the swarm \cite{Kapoutsis2017}, significantly restricting the UAVs cooperation.  
	
	The proposed approach is a semi-distributed, model-free algorithm, that shares the same convergence characteristics with the block coordinate descent (BCD) algorithms, capable of optimizing in real-time the monitor positions of the UAVs to increase the overall situational awareness, without utilizing any prior knowledge about the environment or vehicles' dynamics. Moreover, no prior knowledge about the ``geolocation'' of information-rich areas is needed either. This decision-making scheme is capable of enabling cooperation between different assets towards the accomplishment of a high-level objective. 
	
	The evaluation procedure was designed to avoid all kind of assumptions regarding the detection process. For this reason, in all stages of the development, we directly utilized raw rgb and depth images to actually retrieve the detected objects. AirSim simulator \cite{airsim2017fsr} was utilized as an evaluation platform for its high-fidelity simulations and real-world dynamics. YOLOv3 detector \cite{yolov3} was adopted to leverage from its ability to perform fast detections. An important aspect of this work is that it is modular with respect to the underlying system that detects the objects of interest. Therefore, one could utilize a different detector (tailored to the application needs) and our methodology will still be capable of delivering an optimized set of UAVs' monitor positions, adapting to the detector's specific characteristics.
	
	Fig. \ref{fig:introFig} portraits an indicative example of such a setup. The left-hand side of Fig. \ref{fig:introFig} illustrates the initial monitoring positions for the swam of 4 UAVs. The top, global view of the operational area depicts the fields of view (colored point-clouds) of the all the UAVs, moments after their take-offs. The 4 images below are the current views from their cameras with all the detected objects. The right-hand side of this figure illustrates the converged configuration for the swarm of UAVs. A direct outcome, that can be derived sorely from the bird's-eye view of the operational area, is that the proposed algorithm spread the UAVs taking into consideration the combined field of view from all the point-clouds. However, the most interesting insights can be derived by combining also the 4 images from the UAVs' cameras. All 4 UAVs have been placed in close vicinity of a crossroad, to exploit the fact that these spots usually have the biggest flow of cars/people. A close examination of the images from the vehicles can reveal some extra traits. With the converged positions, the UAVs can monitor multiple road-parts from a single monitoring spot. On top of that, in some cases, with the current orientation of their cameras, the UAVs can monitor roads, that, otherwise, would be physically ``blocked'' by the surrounding buildings/structures (views from UAV2 \& 4). It should be highlighted that none of these features were specifically designed or dictated, but, rather the proposed algorithm found these ``hacks'' to maximize the number and quality of detected objects. 
	
	%The remainder of the paper is structured as follows. Section \ref{sec:setup} defines the structural units of such a multi-UAV framework, to be able to translate the overall mission objectives to an optimization problem. The description of the proposed navigation algorithm, which tackles such a problem, is presented in section \ref{sec:algorithm}. A series of experiments is conducted to adequately analyze the performance of the designed navigation algorithm in section \ref{sec:simulations}. The overall conclusions of the paper are drawn in section \ref{sec:conclusions}.
	
	\section{PROBLEM FORMULATION} 
	\label{sec:setup}
	
	\subsection{Controllable Variables}
	Consider a team (swarm) of $n$ multirotor UAVs $\left\lbrace 1,2,\dots,n\right\rbrace$ cooperate to increase the situational awareness over an unknown, unstructured environment. At each time-step, each UAV is allowed to configure its own monitoring pose by changing its own position and orientation. To be compliant with various security regulations, each UAV is deployed in a distinct, constantly-fixed height level from the ground. Therefore, the controllable variables for each UAV include the position in $xy$ plane and its orientation (yaw). Overall, the decision vector $x_i(k) = [x, y, \mbox{\textit{yaw}}] \in \mathbb{R}^2 \times \mathbb{S}^1$, where $\mathbb{S}^1$ indicates the 1-sphere, will denote the decision variables of $i$th UAV at $k$th time-step. The augmented decision vector for the swarm will be given by the following equation:
	\begin{equation}
	\label{eq:stateSpace}
	\textbf{x}(k) \coloneqq \left[x_1^\top(k), x_2^\top(k),\dots, x_n^\top(k) \right]^\top 
	\end{equation}
	The values of the decision vector cannot be set arbitrary, but they should satisfy a set of operational constraints. More specifically:
	\begin{enumerate}
		\item the change in controllable variables has a maximum value, i.e. assuming the following dynamics:
		$ x_i(k+1) = x_i(k)+u(k+1) $, then $
		\left|  u \right|  \leq u_{max}$ 
		, where $u_{max}$ depicts the moving capabilities of the $i$th UAV. 
		\item the UAVs should remain within the operational area boundaries
		\item the UAVs should avoid collisions with all the stationary (e.g. terrain morphology) and moving obstacles (e.g. other UAVs). 
	\end{enumerate}
	
	The previously defined constraints can be, in general, represented as a system of inequalities:
	\begin{equation}
	\label{eq:constraints}
	{\cal C} \left( \textbf{x}(k) \right)  \leq 0
	\end{equation}
	
	where ${\cal C}$ is a set of nonlinear functions of the decision variables $\textbf{x}(k)$. The analytical form of this function cannot be known in case of operation inside an unknown/dynamically-changing environment; however, from the images coming from the mounted cameras, it can be determined whether a next (in close vicinity) configuration of decision variables $\textbf{x}(k)$ satisfies or violates the set of constraints (\ref{eq:constraints}).
	
	\subsection{Measurements}
	After applying the new decision variable $x_i(k)$, each UAV will be able to perceive a part of the operational area. This perception includes both an RGB $I_i^{\mbox{RGB}}(k)$ and a depth $I_i^{\mbox{DEPTH}}(k)$ image\footnote{It is worth highlighting that, for both the designing of the navigation algorithm (section \ref{sec:algorithm}) and evaluation procedure (section \ref{sec:simulations}), we made no assumptions regarding the sensors' dynamics, rather we included the received raw images from the cameras of UAVs.}. 
	
	As next step, each $I_i^{\mbox{RGB}}(k)$ is fed to a pre-trained Convolutional Neural Network (CNN) \cite{yolov3}, having as objective to detect objects of interest along with the corresponding confidence levels. The center of mass, in each one of these detected objects, is extracted by utilizing the formation of the pixels that lie inside the bounding box of that object. Combining also the information from the depth image $I_i^{\mbox{DEPTH}}(k)$, the dimension of center of mass is expanded to 3D space. The measurement vector $y_i(k)$ for each $i$th UAV contains this list of 3D centers of masses, along with the corresponding confidence levels and the detected class labels. The swarm measurements' vector is comprised as follows:
	\begin{equation}
	\label{eq:measurements}
	\textbf{y}(k) \coloneqq \left[y_1^\top(k), y_2^\top(k),\dots,y_n^\top(k) \right]^\top 
	\end{equation}

	\subsection{Objective Function}
	\label{subsec:objectiveFunction}

	Having projected all these objects in a global 3D frame, we can now retain only the unique objects, by checking the distance between the detected objects, in this common 3D frame. In the case where an object is identified by more than one UAV, the object with less confidence in its detection is discarded.  By doing so, we want, on one hand, to ``force'' UAVs to detect unique objects, and, on the other hand, to reward UAVs that have achieved to detect objects with a high level of certainty.  %For example, in ``\textit{Combined Perception}'' of figure \ref{fig:ObjectiveFunctionDiagram} the black car was detected by both UAVs, however, we want to keep the detection only from the UAV1.
	
	The set $B$ of all uniquely detected objects can be defined as:	$B \coloneqq A \setminus D $, where $D$ denotes the set of all the objects detected more than once. Having calculated the set of unique detected objects $B$, we now proceed to the calculation of  matrix $c$: %which contains $n\times l$ elements, where $l$ is the cardinality of set $B$. The elements of matrix $C$ will take their values as follows: 
	
	\begin{equation}
	\label{eq:confidenceLevel}
	c_{ij} = \left\lbrace \begin{array}{lr}
	0  & \mbox{if not detected by the $i$th UAV} \\
	cl & \mbox{otherwise}
	\end{array} \right. 
	\end{equation}
	
	where $cl$ denotes the \textit{confidence level} of the $j$ unique object from set $B$, as detected by $i$th UAV.
	
	In a nutshell, the mission objectives of the swarm are expressed by the following equation:
	\begin{equation}
	\label{eq:non_smooth_objectiveFunction}
	\bar{{\cal J}}\left(\textbf{x}(k) \right)  = \sum_{j=1}^{m} \max_{i \in \{1,2,\dots,n\}} c_{ij}  
	\end{equation} 
	
	Using standard results from approximation theory \cite{bertsekas1975nondifferentiable}, we introduce a twice differentiable function:
	\begin{equation}
	\label{eq:objectiveFunction}
	{\cal J}\left(\textbf{x}(k) \right) \mbox{ of class } {C}^{1}
	\end{equation}
	that can approximate with arbitrary accuracy the discontinuous $\bar{{\cal J}}\left(\textbf{x}(k) \right)$\footnote{It is worth stressing that this objective function is only utilized to analyze the convergence of the proposed algorithm, therefore, its form is not needed in the implementation procedure. Apparently, any optimum of $\bar{{\cal J}}\left(\textbf{x}(k) \right)$ is also an optimum in ${\cal J}\left(\textbf{x}(k) \right) $.}.
	
	By maximizing of the above objective function, the swarm is ``forced'' to form configurations that not only detect as many unique objects as possible, but also to achieve the highest possible values of the \textit{confidence levels} and, therefore, to increase the overall situational awareness. %Another important aspect here is that the above formulation forms a trade-off between the number of unique detected objects and their confidence levels. If a UAV chooses to move closer to improve the confidence levels of its detections, it may lose completely some of its detected objects. Therefore, if needed, one can ``tweak'' equation (\ref{eq:non_smooth_objectiveFunction}) to give more or less priority in quantity over the quality of objects.
	
	\subsection{Optimization Problem}  
	Given the mathematical description presented above, the problem of choosing the decision variables online for a multi-UAV system, so as to increase the quantity and the quality of the uniquely detected objects, can be described as the following constrained optimization problem:
	\begin{equation}
	\label{eq:OptimizationProblem}
	\begin{array}{rl}
	\mbox{maximize} & (\ref{eq:objectiveFunction})\\
	\mbox{subject to} & (\ref{eq:constraints}) \,
	\end{array}
	\end{equation}
	
	The above optimization problem (\ref{eq:OptimizationProblem}) cannot be tackled with traditional gradient-based algorithms \cite{nesterov2007gradient}, mainly because the explicit forms for the functions ${\cal J}$ and ${\cal C}$ are not available. To make matters worse, jointly optimizing a function over multiple UAVs ($n$), each of which with multiple decision variables, can incur excessively high computational cost.
	
	\begin{figure*}[!th]
		\centering
		\includegraphics[width=0.94\textwidth]{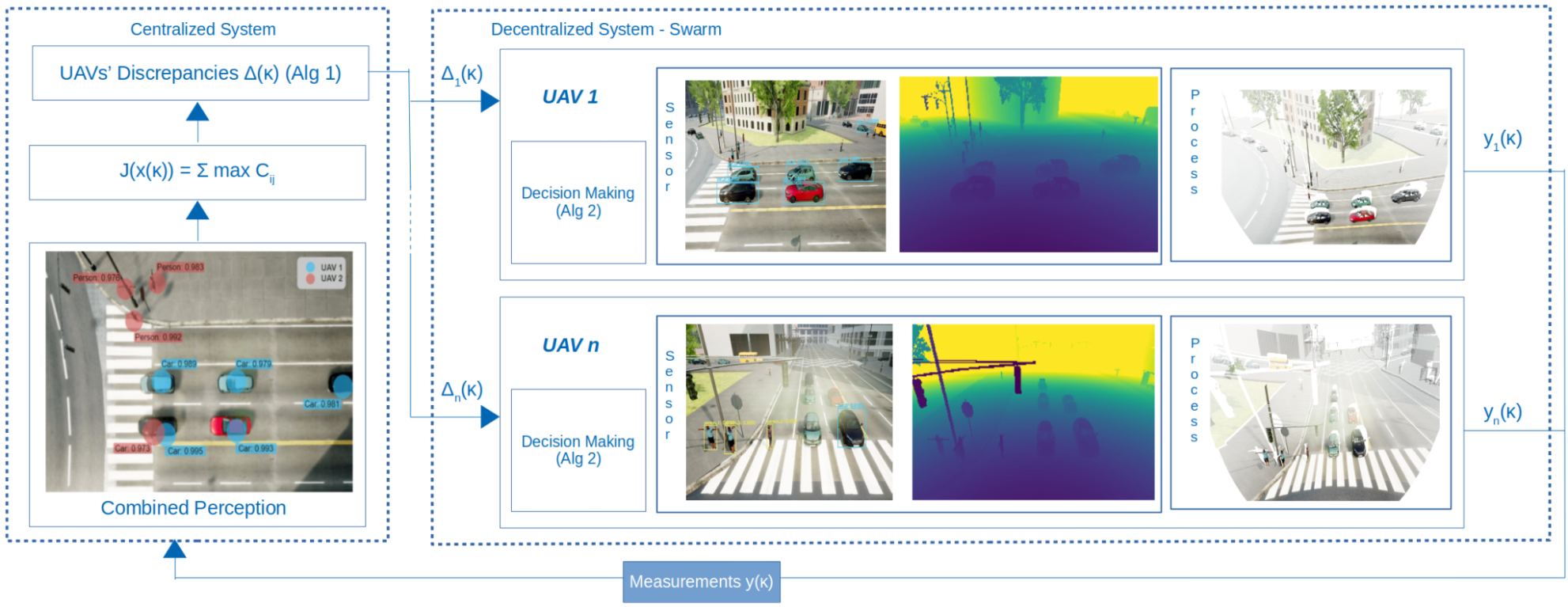}
		\caption{Flowchart of the overall navigation scheme}%Initially each UAV will apply its decision variable $x_i(k)$, based on the Decision Making Algorithm \ref{alg:distributedAlg}. At the new position it will enable its sensor to perceive a part of the operational area ($I_i^{\mbox{RGB}}(k)$ $I_i^{\mbox{DEPTH}}(k)$), it will fuse all of the available information in order to perform the object detection, calculate the center of mass of each detected object and finally it will transmit each measurements yi(k). All of the aforementioned measurements $y_i^\top(k)$ will be combined in the centralized system (Note: the bird's-eye view here is just for illustrative reasons,raw images are not transmitted). The objective function will be calculated/constructed according to (\ref{eq:non_smooth_objectiveFunction}), Algorithm \ref{alg:centralizedAlg} will calculate each UAV's contribution $\Delta_i(k)$ and transmit it back to each UAV respectively.}
		\label{fig:flowchart}
	\end{figure*}
	
	\section{NAVIGATION ALGORITHM}
	\label{sec:algorithm}
	
	To appropriately tackle the aforementioned challenges, and to be able to solve the optimization problem of (\ref{eq:OptimizationProblem}), the following navigation algorithm is proposed. An outline of the proposed navigation scheme is illustrated in Fig. \ref{fig:flowchart}. Overall, the procedure is divided into two distinct parts:
	
	%In this section is presented a navigation algorithm capable of solving the optimization problem as formulated in (\ref{eq:OptimizationProblem}), i.e. to fine-tune the monitoring positions for the team of UAVs to maximize the effective/useful information in their combined field of view.  
	
	\subsection{Global coordination}
	\label{subsec:globalCoordination}
	
	Algorithm \ref{alg:centralizedAlg} outlines the pseudo-code of this first part. Conceptually, this algorithm calculates the contribution of each UAV to the accomplishment of the overall mission objectives (\ref{eq:non_smooth_objectiveFunction}). 
	
	\begin{algorithm}[h]
		\caption{Calculate UAVs' contributions}
		\label{alg:centralizedAlg}
		\hspace*{\algorithmicindent} \textbf{Input:} $\textbf{x}(k), \textbf{y}(k)$\\
		\hspace*{\algorithmicindent} \textbf{Output:} $\Delta(k)$
		\begin{algorithmic}[1]
			\State $\jmath_{k} \gets {\cal J}(\textbf{x}(k))$ (\ref{eq:non_smooth_objectiveFunction}) \Comment{Update system's objective function}
			\ForEach {$i \in \left\lbrace 1,2,\dots,n \right\rbrace  $}
			\State $$\Delta_i(k) \gets \frac{\jmath_{k} - {\cal J}\bigg(\left[ \dots,x_{i-1}(k),x_i(k-1),x_{i+1}(k)\dots\right] \bigg)}{ \left\| x_i(k) - x_i(k-1) \right\| } \;\;$$
			\Comment{Approximation of $ \frac{\Delta x_i^\top}{\left\| \Delta x_i \right\| } \frac{ \partial {\cal J}(k)}{\partial x_i}$}
			\EndFor
		\end{algorithmic}
	\end{algorithm}
	
	%\vartheta
	In line 1, and following the computation scheme of section \ref{subsec:objectiveFunction}, the global objective function index $\jmath_{k}$ is calculated utilizing all the collected images from all the UAVs at the $k$ time-step. Then, for each UAV, we calculate again the objective function, but only this time for the $i$th UAV are used the measurements from the previous time-step ${\cal J}\bigg(\left[ \dots,x_{i-1}(k),x_i(k-1),x_{i+1}(k)\dots\right] \bigg)$. Although this would be a ``new'' UAVs' configuration $\left[\dots, x_{i-1}^\top(k), x_i^\top(k-1), x_{i+1}^\top(k),\dots \right] $, no new movements are required, as all the images from this configuration have already be acquired from the $k$ and $k-1$ time-steps. Finally, line 3 presents the calculation of $\Delta_i(k)$, which is the result from subtracting this previously calculated term from the system's objective function index $\jmath_{k}$, divided by the norm of variation in the decision variables $\left\| x_i(k) - x_i(k-1) \right\|$ (where $\left\| \cdot \right\|$ denotes the Euclidean distance). All in all, $\Delta_i(k)$ carries information of  $\frac{\partial {\cal J}(k)}{\partial x_i}$, which, of course, cannot be derived analytically, as the algebraic form that relates the monitoring positions of the UAVs (\ref{eq:stateSpace}) and the objective function (\ref{eq:non_smooth_objectiveFunction}) is environment/dynamics-dependent and generally not available. 
	
	\subsection{Distributed decision}
	\label{subsec:distributedDecision}
	
	Each $\Delta_i(k)$ is dispatched to the $i$th UAV, and, from this step and after, all calculations are performed locally, building a system that i) is resilient to UAV failures, ii) does not require any other global coordination, and ii) all the decision variables' updates are made in a (parallel) distributed fashion. The pseudo-code for the update on the monitoring position of each UAV based on the $\Delta_i(k)$ is given in Algorithm \ref{alg:distributedAlg}.
	
	\begin{algorithm}[h]
		\caption{Decision making on the $i$th UAV}
		\label{alg:distributedAlg}
		\hspace*{\algorithmicindent} \textbf{Input:}$\Delta_i(k)$\\
		\hspace*{\algorithmicindent} \textbf{Output:} $x_i^\top(k+1)$\\
		\hspace*{\algorithmicindent} \textbf{Hyperparameters:} $\phi_i$, $T(k)$, $m$, and $\alpha(k)$
		\begin{algorithmic}[1]
			\State $J_i(k) \gets J_i(k-1) + \Delta_i(k)$ \Comment{Local optimization}
			\State $J_i(k+1) \approx \hat{J}_i(k+1) = \theta_i^\top(k)\phi_i\big(x_i(k)\big)$  \Comment{Construct a LIP polynomial estimator}
			\State $\theta_i^* \gets \argmin_\theta \sum_{\ell = k-T(k)}^{k} \bigg( \theta^\top \phi_i\big(x_i(\ell)\big) - J_i(\ell)\bigg)^2 \;\;\;\;$ \Comment{Solve least-squares optimization problem}
			\State $\delta x_i^{(1)}(k),\delta x_i^{(2)}(k),\dots,\delta x_i^{(m)}(k)$ \Comment{Generate randomly a set of $m$ candidate perturbations}
			\State exclude the ones that violate (\ref{eq:constraints})
			\State $j^* \gets \argmax_{j=1,\dots,m} \theta_i^{*\top}\phi_i\bigg(x_i(k) + \alpha(k)\delta x_i^{(j)}(k)\bigg)$ \Comment{Find the perturbation that maximizes the previously constructed estimator}
			\State $x_i(k+1) \gets x_i(k) + \alpha(k)\delta x_i^{(j^*)}(k)$ \Comment{Update the decision variables of the $i$th UAV}
		\end{algorithmic}
	\end{algorithm}
	
	In line 1, we update the value of the local objective function $J_i(k)$ to evaluate the previously executed command $x_i(k)$. $ J_i(k)$ encapsulates only the elements that affect the performance of (\ref{eq:non_smooth_objectiveFunction}) for the $i$th UAV, considering the monitoring positions of all the other UAVs as part of the problem to be solved (system's dynamics). In lines 2 \& 3, we construct a Linear-In-the-Parameters estimator, $T(k)$ denotes the variable length of estimation window, to approximate the evolution of the $i$th sub-cost function $J_i(k)$. As a next step (lines 4-5), we generate randomly a set of $m$ perturbations $\delta x_i$, around the current monitoring position, respecting the operational constraints (\ref{eq:constraints}). Having built an estimator for $J_i(k)$, we are now in the position to evaluate numerically -- without the need to physically execute the movement -- all these valid next candidate monitoring positions, $x_i^j(k) = x_i(k) + \alpha(k) \delta x_i^j(k), \; \forall j \in \{1,\dots,m\}$ (line 6)\footnote{$\alpha(k)$ is a positive function chosen to be either a constant positive function or a time-descending function satisfying $\alpha(k) >0, \sum_{k=0}^\infty \alpha(k) = \infty, \sum_{k=0}^\infty \alpha(k)^2 < \infty$. For more details regarding the form and tuning of $\phi_i$, $T(k)$, $m$, and $\alpha(k)$ please check \cite{kapoutsis2019distributed,kosmatopoulos2009adaptive,kosmatopoulos2009large}}. The updated monitoring position $x_i(k+1)$ is the candidate that maximizes the previously constructed estimator $\hat{J}_i$ (line 7).
	  
%	\footnote{Furthermore, $\alpha(k) \leq \bar{\alpha} \; \forall k$, where $\bar{\alpha}$ is a problem-specific constant, correlated with the UAV's dynamics and the mission objectives (\ref{eq:non_smooth_objectiveFunction}}  
	%\footnote{As shown in \cite{kosmatopoulos2009adaptive, kosmatopoulos2009large}, it is sufficient to choose the number perturbations $m$ to be greater than $2 \times n$.}  
	  
	\subsection{Convergence}
	\label{subsec:converagence}
	
	In a nutshell, at each time-step, Algorithm \ref{alg:centralizedAlg} is applied to calculate the contribution vector $\Delta(k)$ and then, in a distributed fashion, Algorithm \ref{alg:distributedAlg} is applied, as many times as the number of UAVs, to update the decision variables $x_i(k+1),\; \forall i\in\{1,2,\dots,n\}$.
	
\begin{remark} 
	\label{rm:BCD}
	From the definition of the term $\Delta_i(k)$, we have that 
	\begin{equation}
	\label{eq:di_fraction}
		\Delta_i(k) = \frac{ {\cal J}(\textbf{x}(k)) - {\cal J}\left( \ldots, x_{i-1}(k), x_i(k-1), x_{i+1}(k), \ldots\right) }{\left\| \Delta x_i(k) \right\| }
	\end{equation}
	where ${\cal J}(\textbf{x}(k)) =  {\cal J}\left( \ldots, x_{i-1}(k), x_i(k), x_{i+1}(k)\ldots \right)$ and $\Delta x_i(k)=x_i(k) - x_i(k-1)$. Thus, by employing Taylor approximation arguments, see e.g. {\em Lemma 7} of \cite{kosmatopoulos2009adaptive}, we have that
	\begin{equation}
	\label{eq:di}
		\Delta_i (k)= \frac{\Delta x_i^\top(k)}{\left\| \Delta x_i (k)\right\| } \frac{ \partial {\cal J}(k)}{\partial x_i}(k) +\mathcal{O}\left(\left\| \Delta x_i(k) \right\| \right)  
	\end{equation}
	where the term $\mathcal{O}\left(\left\| \Delta x_i(k)) \right\| \right) $ stands for the error introduced, due to the fact the difference between two consecutive monitoring positions is not negligible. Therefore as the difference between the $x_i(k-1)$ and  $x_i(k)$ gets smaller (definition of $\alpha(k)$), $\Delta_i$ will better approximate the term  $\frac{\Delta x_i^\top(k)}{\left\| \Delta x_i (k)\right\| } \frac{ \partial {\cal J}(k)}{\partial x_i}(k)$.
\end{remark}

\begin{remark} 
	\label{rm:convergenceOfCAO}
	As shown in \cite{kosmatopoulos2009adaptive, kosmatopoulos2009large}, the decision-making scheme implemented in each UAV (Algorithm \ref{alg:distributedAlg}) guarantees that If $m \geq 2 \times \text{dim}\left(x_i\right)$, the vector $\phi$ satisfies the universal approximation property \cite{hassoun1995fundamentals} and the functions $J_i$ and $C$ are  continuous with continuous first derivative, then the update rule of $x_i$ (Algorithm \ref{alg:distributedAlg}, line 7) is equivalent to
	\begin{equation}
	\label{eq:dxi}
	\Delta x_i (k) =   \gamma(k) \frac{ \partial {\cal J}(k)}{\partial x_i}(k) + \epsilon(k)
	\end{equation}
	where $\gamma(k)$ is a positive scalar quantity that depends on the choice of $\alpha$.
	 As it was shown in \cite{kosmatopoulos2009adaptive, kosmatopoulos2009large} the term $\epsilon(k)$ equals $ \alpha(k) \times \nu$, where $\nu$ being an approximation error due to presence of LIP polynomial estimator. Since $\alpha(k)$ converges to zero, $\epsilon(k)$ will be vanished as well.
\end{remark}

	\begin{theorem}
		\label{th:1}    
		The local convergence of the proposed algorithm can be guaranteed in the general case where the system's cost function ${\cal J}$ and each UAV's contribution $J_i$ are non-convex, non-smooth functions.
	\end{theorem}
	
	\textit{sketch of the proof:} Substituting equation \ref{eq:dxi} to \ref{eq:di} we obtain 
\begin{equation}
	\label{eq:proof_1}
		\Delta_i (k)=  \frac{\gamma(k)}{\left\| \Delta x_i (k)\right\| } \left\|\frac{\partial {\cal J}(k)}{\partial x_i}(k) \right\|+\mathcal{O}\left(\left\| \Delta x_i(k) \right\| \right) \bar{\epsilon}(k)
	\end{equation}
	where $\bar{\epsilon}(k)$ is an ${\cal O}(\epsilon(k))$ term and, thus, it converges asymptotically to zero. From equations \ref{eq:proof_1} and \ref{eq:di_fraction} and the fact that $\lim_{t\mapsto \infty} \Delta x_i(k) =0$ (due to the use of the term  $\alpha(k)$), we can establish that Algorithm  \ref{alg:distributedAlg} guarantees convergence to a local optimum of  $J_i$, i.e., that it guarantees that $x_i$ will converge to its locally optimum value, that satisfies
	$$
	\frac{\partial {\cal J}\left(x_1,\dots,x_{i-1},w,x_{i+1},\dots,x_{n}\right)}{\partial w} = 0
	$$
	subject to (\ref{eq:constraints}) and therefore, the proposed algorithm approximates the behavior of the block coordinate descent (BCD) \cite[Algorithm 1]{wright2015coordinate} family of approaches. Following the proof described in \cite[Proposition 2.7.1]{bertsekas1999nonlinear}, it is straightforward to see that if the maximum with respect to each block of variables is unique, then any accumulation point of the sequence $\{\textbf{x}(k)\}$ generated by the BCD methodology is also a stationary point.

	%\textit{sketch of the proof:} Remark \ref{rm:BCD} establishes that indeed Algorithm \ref{alg:centralizedAlg} provides approximation of $\frac{\partial {\cal J}}{\partial x_i}, \; \forall i \in\left\lbrace 1,2,\dots,n \right\rbrace $ and Remark \ref{rm:convergenceOfCAO} establishes the Algorithm  \ref{alg:distributedAlg} is going to convergence to a local optimum of the true value of $J_i$. Combining these two outcomes, the distributed update on each UAV is equivalent to
	%$$
	%x_i = \argmax_{w} {\cal J}\left(x_1,\dots,x_{i-1},w,x_{i+1},\dots,x_{n}\right)  \\
	%$$
	%subject to (\ref{eq:constraints}) and therefore, the proposed algorithm approximates the behavior of the block coordinate descent (BCD) \cite[Algorithm 1]{wright2015coordinate} family of approaches. Following the proof described in \cite[Proposition 2.7.1]{bertsekas1999nonlinear}, it is straightforward to see that if the maximum with respect to each block of variables is unique, then any accumulation point of the sequence $\{x(k)\}$ generated by the BCD methodology is also a stationary point. 
	
	\section{EVALUATION RESULTS}
	\label{sec:simulations}
	
	\subsection{Implementation details}
	AirSim platform \cite{airsim2017fsr} was utilized to evaluate the proposed algorithm. AirSim is an open-source, cross-platform, simulator for drones, cars and more that supports hardware-in-loop with popular flight controllers for physically and visually realistic simulations. All the experiments were carried out in the \textit{CityEnv} of the aforementioned simulator, which is a vast, realistic environment that simulates both the static structures and the highly non-linear and dynamic behavior of the moving assets (e.g. cars, trucks, pedestrians, etc.). %However, this spatial constraint restricts only the positions of UAVs and not the detected objects. The choice of such a spot was made by taking into consideration both the extensive complexity of the road network and also the unpredictable variations in the traffic flow. 
	
	The swarm consists of simulated multirotor UAVs instantiated with the AirSim built-in flight controller (called \textit{simple\_flight}) and equipped with stationary cameras. Each stationary camera is located at the front and rotated downwards $45^o$ on the pitch axis. Each UAV can move inside the $xy$ plane of the previously defined operational area and also has a $360^o$ yaw movement, i.e. the controllable variables as defined in (\ref{eq:stateSpace}). The operational height of the first UAV is at $14m$ from the ground, while each additional UAV is deployed at a height $0.5m$ higher than the previous one. The maximum step-size (Algorithm \ref{alg:distributedAlg}, line 7) was set to be $3.5m$ for the movement inside $xy$ plane and $10^o$ for the yaw rotation.
	
	The calculation of detected objects, in each one of the received RGB images of the UAVs (subsection \ref{subsec:objectiveFunction}), was achieved by employing YOLOv3 \cite{yolov3} detector pretrained on COCO dataset \cite{lin2014microsoft}. YOLOv3 is indeed a state-of-the-art object detector capable of producing reliable predictions in real-time, however, it has not been trained for data coming from a simulator, nor for top-view images (typical UAV's images.). Although this unsuitability of the object detector caused several problems (e.g. objects completely missed if they were observed from different angles, radical changes in the confidence levels of detected objects caused by slight changes in the pose of the UAV, etc.) in the evaluation of the UAVs' configuration, we chose to apply the object detector ``as is'', to highlight that the proposed navigation algorithm does not utilize any information related to these choices, therefore it is modular with respect to alternative systems.
	
	The hyperparameters and variables of algorithm \ref{alg:distributedAlg} chosen so as: $i)$ $\phi_i$ to be a 3rd degree polynomial, $ii)$ constant estimator's time winows of $T(k) = 30 \; \forall k$, $iii)$ $m=40$ perturbations to be evaluated on the estimator, and, $iv)$  $\alpha(k) = 1.1 - \frac{k}{300}$. 
	
	\subsection{Performance on different operational areas}
	\label{subsec:areas}
\begin{figure}[!t]
		\centering
		\begin{subfigure}[b]{0.165\textwidth}
			\centering
			\includegraphics[width=1\linewidth]{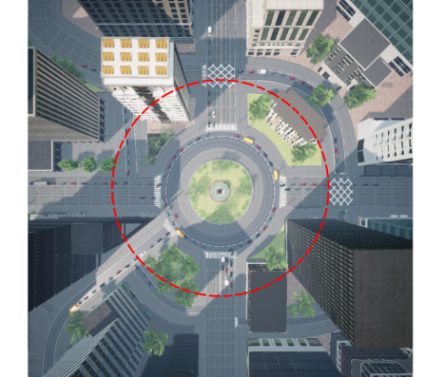}
			\caption{Area $\#1$}
			\label{fig:area1}
		\end{subfigure}%		
		\begin{subfigure}[b]{0.165\textwidth}
			\centering
			\includegraphics[width=1\linewidth]{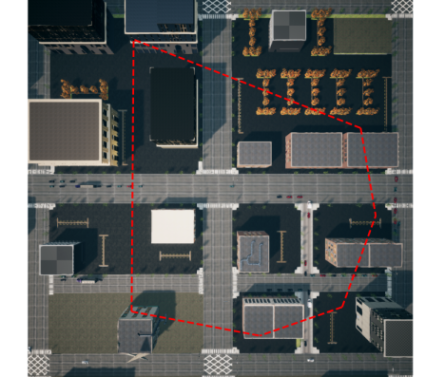}
			\caption{Area $\#2$}
			\label{fig:area2}
		\end{subfigure}%
		\begin{subfigure}[b]{0.165\textwidth}
			\centering
			\includegraphics[width=1\linewidth]{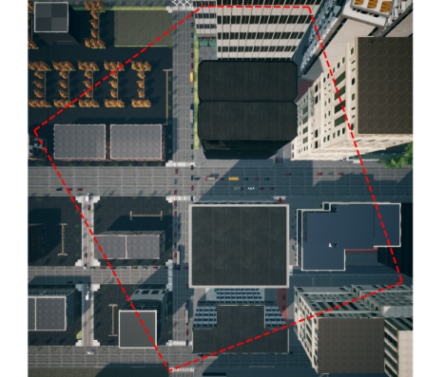}
			\caption{Area $\#3$ }
			\label{fig:area3}
		\end{subfigure}
		\caption{Operational areas: (a) Airsim central roundabout, (b) Office complex neighborhood, (c) Semi-urban region}
		\label{fig:areas}
	\end{figure}

	Initially, we investigate the performance of the proposed navigation algorithm in 3 different terrains as depicted in Fig. \ref{fig:areas}. The first operational area (Fig. \ref{fig:area1}) is defined on top of the ``central'' roundabout of the environment. The UAVs were allowed to move inside a circle centered at the center of the roundabout and with a radius of $70m$, covering an area of $15393.8m^2$. Area $\#2$ (Fig. \ref{fig:area2}) is defined as a polygon above an office complex with a total area of $37892.3m^2$. Finally, area $\#3$ (Fig. \ref{fig:area2}) was selected to be on top of a semi-urban area having a total area of $56894.9m^2$.
	
\begin{figure}[!t]
		\centering
		\includegraphics[width=0.97\columnwidth]{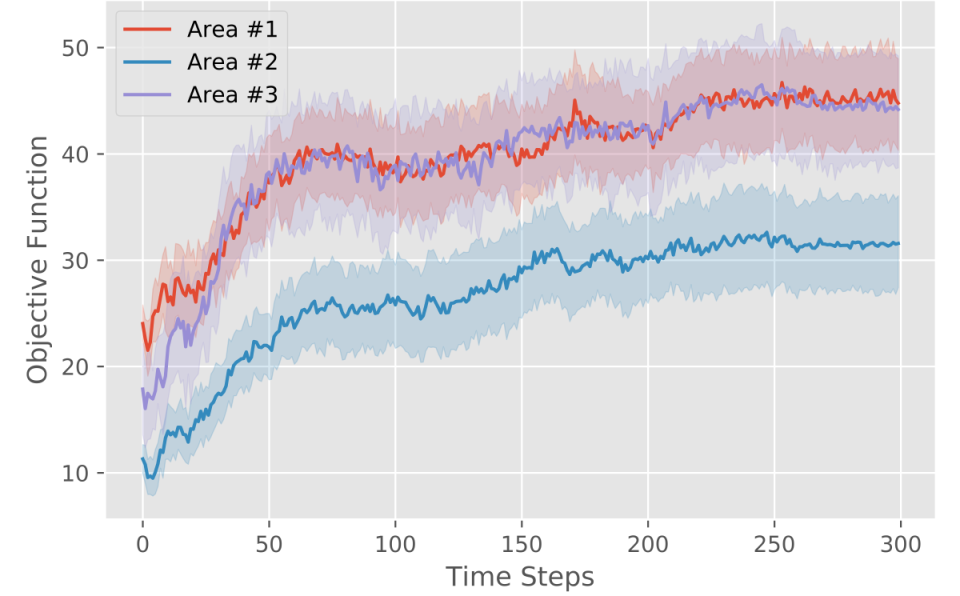}
		\caption{Convergence analysis on the terrains of figure \ref{fig:areas}}
		\label{fig:averageAreas}
	\end{figure}

	In each one of the above described areas we performed 40 experiments with randomly selected starting positions for a swarm of 4 UAVs. The performance for each one of these areas is depicted in Fig. \ref{fig:averageAreas}. A direct comparison between the results on these different areas is not valid as the size, geometry of the operational area and the structure of road network are not the same. Nevertheless, by examining the results in these divergent set-ups, it is evidently that the proposed navigation algorithm is able to guide the swarm to a acceptable configuration extremely fast, i.e. within only 50 time-steps. Also, in the majority of the cases it had converged to the final configuration around the time-step 200. In the remaining subsections we focus the experimental validation on the first area (Fig. \ref{fig:area1}), presenting an in-depth performance analysis.
	
	\subsection{Performance Comparison With Centralized Semi-Exhaustive Search}
	\label{subsec:ExhaustiveSearch}
	
	Before continuing with the analysis of the evaluation results, let us define an algorithm that has little practical value, however, its achieved performance can provide us with valuable insights, when compared with the proposed navigation algorithm. This algorithm is a centralized, semi-exhaustive methodology that works as follows: At each time-step, first, it generates a subset (\textit{semi-exhaustive}) of candidate UAVs' configurations (\textit{centralized}) out of all possible ones. Then, all these candidates are evaluated on the AirSim platform, i.e. the UAVs have to actually reach that candidate monitoring positions, and, for each one of them, is calculated also the objective function (following the computation scheme from subsection \ref{subsec:objectiveFunction}). Finally, the next configuration for the swarm is the candidate maximizes the objective function value. This procedure is repeated for every time-step of the experiment. 
	
	As the number of candidate UAVs' configurations (that are evaluated) increases, the performance of this algorithm will approximate the best possible one. The biggest asset of this approach, to be able to evaluate configurations of UAVs before deciding about their next movement, is also its major disadvantage, as it renders this decision-making scheme unfeasible and unsuitable for any kind of real-life applications. 
	
	\begin{figure}[!t]
		\centering
		\includegraphics[width=0.97\columnwidth]{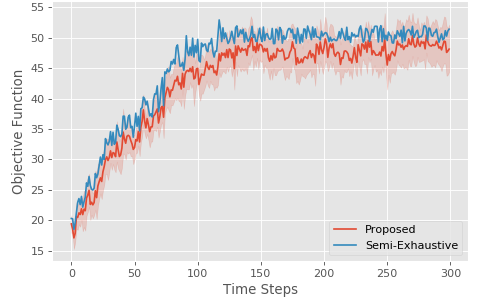}
		\caption{Performance comparison between the proposed algorithm and the centralized, semi-exhaustive search}
		\label{fig:averageJDiagram}
	\end{figure}
	
	Fig. \ref{fig:averageJDiagram} presents a comparison study between the semi-exhaustive search and the proposed algorithm. 4 UAVs were deployed in the same operational area as defined in Fig. \ref{fig:area1}. For both algorithms, the experiment lasted 300 time-steps, where the maximum step-size was as defined in the $2^{\mbox{nd}}$ paragraph of section \ref{sec:simulations}. The semi-exhaustive algorithm evaluated 60 different UAVs' configurations before deciding about the next monitoring position. The proposed algorithm was capable of producing a similar behavior with the semi-exhaustive algorithm, having almost the same convergence rate. The final converged value (average value after 150 time-steps) was $47.64$ for the proposed algorithm and $50.97$ for the semi-exhaustive algorithm. All in all, this analysis highlights that the proposed algorithm is capable of achieving a performance that is equivalent to having the ``luxury'' to spare 60 different combinations of UAVs, before deciding about the next configuration of the swarm. 
	
	\subsection{Areas of Convergence}
	\label{subsec:Convergence AreasTrajecotries}
	%In this subsection, we look deeper into the results of figure \ref{fig:averageJDiagram} for the proposed algorithm, and provide with extra insights and visual results about its operation and robustness.
	
	%\begin{figure}[!th]
	%	\centering
	%	\if\imagesQuality0
	%	\includegraphics[width=0.99\columnwidth]{images/results_5.png}
	%	\else
	%	\includegraphics[width=0.99\columnwidth]{images/results_5_Original.png}
	%	\fi
	%	\caption{Evolution of the trajectories as calculated in real-time by the proposed algorithm}
	%	\label{fig:trajecotries}
	%\end{figure}
	
	%Figure \ref{fig:trajecotries} graphically illustrates the trajectories of all the UAVs from their taking-off positions to their converged ones, for a single instance, as calculated by the proposed algorithm. This visual representation of the resulting trajectories validates the fast convergence of the proposed navigation algorithm, as it was initially derived from figure \ref{fig:averageJDiagram}. The fluctuations of the UAVs around the converged positions are an essential ingredient of Algorithm \ref{alg:distributedAlg}, that allows rapid adaptation of the swarm to cope with changes in the environment (e.g. sudden changes of the traffic flow, congestion, etc.). 
	
	\begin{figure}[!t]
		\centering
		\includegraphics[width=0.97\columnwidth]{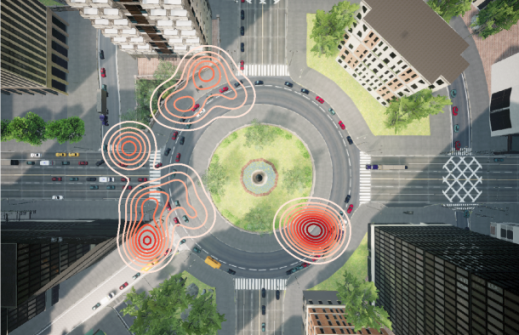}
		\caption{Equipotential lines of final monitoring positions of the UAVs}
		\label{fig:finalDensity}
	\end{figure}
	
	Fig. \ref{fig:finalDensity} visualizes the density distribution of the converged final positions ($100$ last positions $\times$ $40$ experiments) with equipotential  lines, again as derived by the proposed algorithm. A straightforward outcome is that the proposed navigation scheme is consistent with respect to the final monitoring positions, as it forms 4 spatially restricted, distinct clusters (one for each UAV). Examining the locations of these 4 clusters, with respect to the geometry of the operational area, we can see that all these clusters are close to a crossroad, where the traffic is usually higher. Finally, the centers of these clusters are strategically placed to able to detect objects from multiple road-parts, from a single monitoring position with fixed orientation (correlation with Fig. \ref{fig:introFig}).
	
	\subsection{Scale Up Study}
	\label{subsec:SwarmSizeComparison}
		\begin{figure}[t]
		\centering
		\includegraphics[width=0.97\columnwidth]{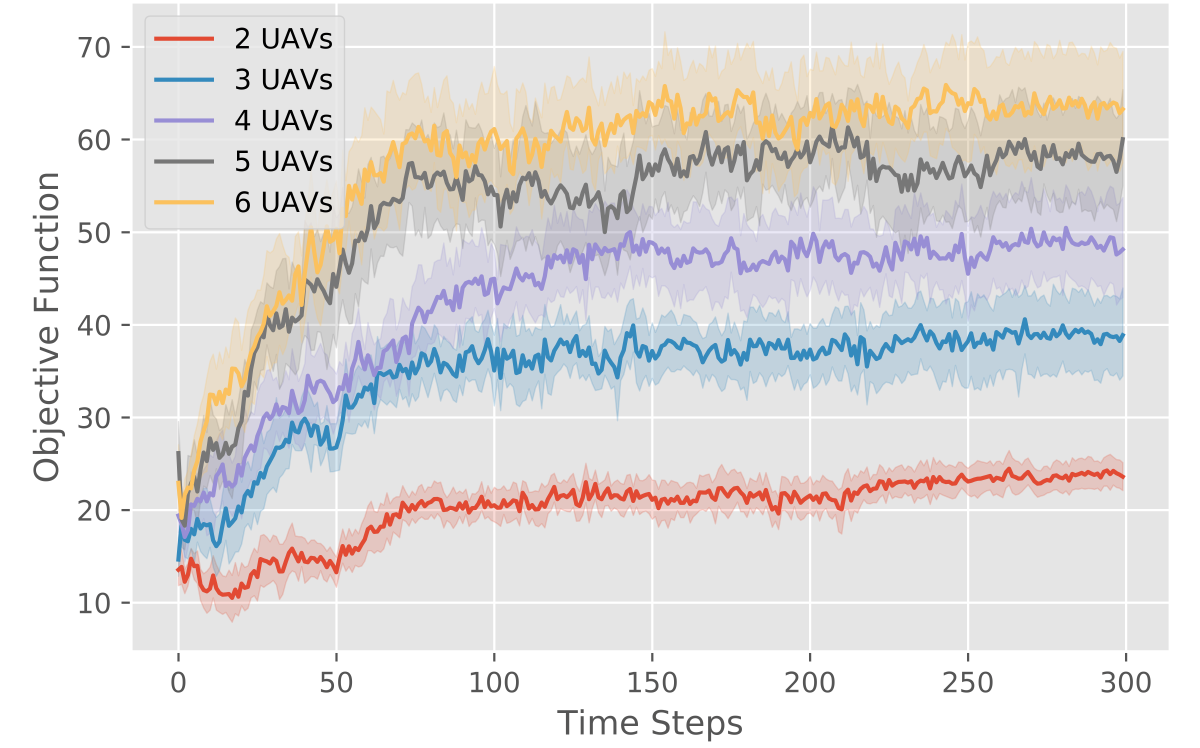}
		\caption{Evolution of the objective function (average \& corresponding standard deviation) for different swarm sizes}
		\label{fig:averageJAll}
	\end{figure}
	
\begin{figure}[!t]
		\centering
		\includegraphics[width=0.98\columnwidth]{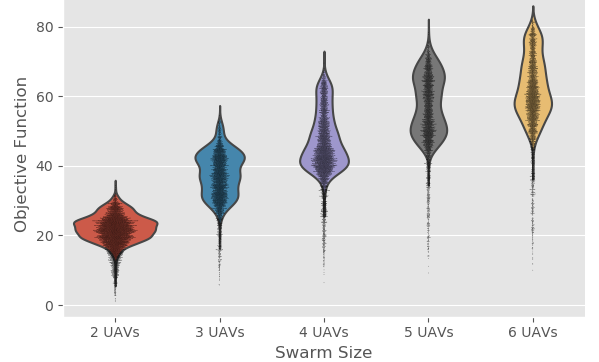}
		\caption{Comparison study on the achieved score for different swarm sizes}
		\label{fig:comparisonDiagram}
	\end{figure}

	We close this analysis by conducting a comparison study on the number of UAVs deployed for the monitoring mission. For this reason, we retained the same simulation environment as before, and we deployed swarms from $2$ and up to $6$ UAVs. To average out the effect of random formation/flow of the traffic in the AirSim platform, for each different configuration, a series of $20$ experiments was performed. Fig. \ref{fig:averageJAll} and \ref{fig:comparisonDiagram} summarize the results of such evaluation. 

	Fig. \ref{fig:averageJAll} depicts the evolution of the objective function over the horizon of the experiment, focusing on the rate of convergence and the deviation around the average performance. For each scenario, the thick colored lines stand for the evolution of the average value, whilst the transparent surfaces around them correspond to the standard deviation. Fig. \ref{fig:comparisonDiagram} is a violin diagram of the last $100$ values (from where no improvement is foreseen, see Fig. \ref{fig:averageJAll}) of the objective function, for each swarm size. The actual values are depicted with the black points and the colored surface around each cluster is a distribution of the density.
	
	Overall, when the number of UAVs inside the swarm is increased, the achieved value of the objective function will increase too. However, this does not mean necessarily that the configurations with higher number of UAVs observe more objects of interest explicitly. It is possible for two configurations of UAVs to monitor more or less the same number of objects of interest. However, in one of them, the redundancy in terms of UAVs, enables the monitoring (detection) with higher confidence levels. This fine-grained evaluation of the monitoring positions of the UAVs is achieved by the designing of the objective function (section \ref{subsec:objectiveFunction}). By examining the performance of each configuration, on one hand, 2 UAVs cannot be informative enough, as only two captures of the environment are insufficient for such a large area. On the other hand, adding more than 5 UAVs, hardly provides a significant improvement, as the overall performance seems to saturate. On top of that, there are several cases, where configurations with 5 UAVs outperformed ones with 6 (Fig. \ref{fig:averageJAll}, deviation around the thick yellow line). The causality behind this phenomenon was the fact that the plethora of UAVs led one (or more) UAV to get ``trapped'' among others, and therefore, not contributing to the overall monitoring task.
	
	\section{CONCLUSIONS}
	\label{sec:conclusions}
	A semi-distributed algorithm for optimizing the monitoring positions for a swarm of UAVs to increase the overall situational awareness has been proposed. A fine-grained analysis of the different components of the algorithm revealed that it shares the same convergence characteristics as those of BCD algorithms. One of the fundamental elements of this proposed algorithm is the fact that it is not specifically tailored to the dynamics of either UAVs or the environment, instead, it learns, from the real-time images, exactly the most effective formations of the swarm for the underlying monitoring task. Moreover, and to be able to evaluate at each iteration the swarm formation, images from the UAVs are fed to a novel computation scheme that assigns a single scalar score, taking into consideration the number and quality of all unique objects of interest. An extensive evaluation testing on the realistic AirSim platform along with a comparison with a semi-exhaustive methodology proved the efficiency of the proposed approach. Future work includes the transition to a completely decentralized algorithm, where each UAV decides about its local controllable variables having as ultimate objective to maximize an estimation of the system's objection function \cite{cortes2008distributed} and to apply the aforementioned methodology in a real-world set-up.

	%\addtolength{\textheight}{0cm}   % This command serves to balance the column lengths
	
	%%%%%%%%%%%%%%%%%%%%%%%%%%%%%%%%%%%%%%%%%%%%%%%%%%%%%%%%%%%%%%%%%%%%%%%%%%%%%%%%

	%%%%%%%%%%%%%%%%%%%%%%%%%%%%%%%%%%%%%%%%%%%%%%%%%%%%%%%%%%%%%%%%%%%%%%%%%%%%%%%%

	%%%%%%%%%%%%%%%%%%%%%%%%%%%%%%%%%%%%%%%%%%%%%%%%%%%%%%%%%%%%%%%%%%%%%%%%%%%%%%%%
	
	\section{ACKNOWLEDGMENT}
	We gratefully acknowledge the support of NVIDIA Corporation with the donation of the Titan V GPU used for this research.

	%%%%%%%%%%%%%%%%%%%%%%%%%%%%%%%%%%%%%%%%%%%%%%%%%%%%%%%%%%%%%%%%%%%%%%%%%%%%%%%%
	\bibliographystyle{IEEEtran}
	\bibliography{references}

\end{document}